\documentclass[runningheads]{llncs}

\usepackage{graphicx}

\usepackage[backend=biber, style=numeric-comp, maxbibnames=99, sorting=none]{biblatex}
\usepackage{multirow}
\usepackage{amssymb}
\usepackage{amsmath}
\usepackage{dsfont}
\usepackage[utf8]{inputenc}
\usepackage[T1]{fontenc}
\usepackage[english]{babel}
\usepackage{mathtools}
\usepackage[ruled]{algorithm2e}
\usepackage{algpseudocode}

\usepackage{subfig}

\usepackage[colorinlistoftodos,prependcaption]{todonotes}
\usepackage{blindtext}
\usepackage{hyperref}

\begin{document}

\title{Adaptable image quality assessment using meta-reinforcement learning of task amenability}
\titlerunning{Meta-RL for image quality assessment}

\author{Shaheer U. Saeed\inst{1} \and
Yunguan Fu\inst{1, 2} \and
Vasilis Stavrinides\inst{3, 4} \and
Zachary M. C. Baum\inst{1} \and
Qianye Yang\inst{1} \and
Mirabela Rusu\inst{5} \and
Richard E. Fan\inst{6} \and
Geoffrey A. Sonn\inst{5, 6} \and
J. Alison Noble\inst{7} \and
Dean C. Barratt\inst{1} \and
Yipeng Hu\inst{1,7}
}
\authorrunning{S.U. Saeed et al.}

\institute{
Centre for Medical Image Computing, Wellcome/EPSRC Centre for Interventional \& Surgical Sciences, and Department of Medical Physics \& Biomedical Engineering, University College London, London, UK \and 
InstaDeep, London, UK \and 
Division of Surgery \& Interventional Science, University College London, London, UK \and 
Department of Urology, University College Hospital NHS Foundation Trust, London, UK \and 
Department of Radiology, Stanford School of Medicine, Stanford, California, USA \and 
Department of Urology, Stanford School of Medicine, Stanford, California, USA \and 
Department of Engineering Science, University of Oxford, Oxford, UK \\ 
\email{shaheer.saeed.17@ucl.ac.uk}
}

\maketitle              

\begin{abstract}

The performance of many medical image analysis tasks are strongly associated with image data quality. When developing modern deep learning algorithms, rather than relying on subjective (human-based) image quality assessment (IQA), task amenability potentially provides an objective measure of task-specific image quality. To predict task amenability, an \textit{IQA agent} is trained using reinforcement learning (RL) with a simultaneously optimised \textit{task predictor}, such as a classification or segmentation neural network. In this work, we develop transfer learning or adaptation strategies to increase the adaptability of both the IQA agent and the task predictor so that they are less dependent on high-quality, expert-labelled training data. The proposed transfer learning strategy re-formulates the original RL problem for task amenability in a meta-reinforcement learning (meta-RL) framework. The resulting algorithm facilitates efficient adaptation of the agent to different definitions of image quality, each with its own Markov decision process environment including different images, labels and an adaptable task predictor. Our work demonstrates that the IQA agents pre-trained on non-expert task labels can be adapted to predict task amenability as defined by expert task labels, using only a small set of expert labels. Using 6644 clinical ultrasound images from 249 prostate cancer patients, our results for image classification and segmentation tasks show that the proposed IQA method can be adapted using data with as few as respective 19.7$\%$ and 29.6$\%$ expert-reviewed consensus labels and still achieve comparable IQA and task performance, which would otherwise require a training dataset with 100$\%$ expert labels.

\keywords{Image quality assessment  \and Meta-reinforcement learning \and Task amenability \and Ultrasound \and Prostate.}

\end{abstract}

\section{Introduction}

Medical image quality can influence the downstream clinical tasks intended for medical images \cite{qa_review}. Automated algorithms have been proposed for image quality assessment (IQA), based on human scoring of image quality \cite{qa_liver_deep, qa_retina_deep, qa_zachary_lung, abdi_qa_ec, liao_qa_ec}, prior clinical knowledge \cite{qa_fetal, lin_qa_multitask} or a set of hand-engineered criteria \cite{qa_retina, qa_vessel, us_qa_carotid}. Task-specific image quality, which measures how well a clinical task can be completed using the image being assessed, may be preferred, but previous methods still rely on human interpretation \cite{qa_liver_deep, qa_retina_deep}. When the downstream clinical tasks are completed by automated machine learning algorithms, task-specific IQA may become more relevant, however, human perceived task-specific IQA may not accurately reflect the performance of the machine optimised \textit{task predictors}. Recent works introduce task amenability; defined as the task-specific image quality to directly measure target task performance \cite{saeed_amenability, google_dvrl}, which also takes into account the dependency between training an automated IQA and the training of a task predictor.

For predicting task amenability for IQA, Saeed \textit{et al.} \cite{saeed_amenability} proposed to train a controller; here, a reinforcement learning (RL) agent, together with the task predictor. Classification and segmentation neural networks were tested as the task predictors. The trained controller predicts significantly different task amenability scores to those determined by humans, with or without requiring human labels of task amenability during training.

By definition, this IQA approach is inevitably dependent on the task predictor and the labelled data used to train such a task predictor, in the case of supervised learning. In clinical practice, the feasibility and cost associated with obtaining quality labelled data sets for various target tasks can not be overlooked. Therefore, we propose a transfer learning strategy to train the IQA agent based on meta-reinforcement learning (meta-RL) across multiple environments. These RL environments can then be designed to reflect different Markov decision processes (MDPs) with differently labelled data. At the same time, a shared task predictor\footnote[1]{\textit{Tasks} refer to the target classification or segmentation tasks, while MDPs or environments are preferred over \textit{meta-tasks} found in meta-learning literature for clarity.} is trained between these MDPs, such that it may be adapted together with the meta-trained controller. Equipping adaptation ability to both the controller and the task predictor has several potential applications for the efficient use of labelled data. In this work, we demonstrate the resulting adaptation ability from relatively low-quality \textit{non-expert} task labels annotated by individual observers to high-quality \textit{expert} labels carefully curated by reviewed consensus.

The contributions of the work are summarised as follows: 1) we propose a transfer learning or adaptation strategy to train an adaptable IQA system; 2) we design a meta-RL algorithm for training the task-amenability-predicting controller together with a target task predictor, which is shared amongst multiple environments, such that training to convergence is not required on every time-step and where adaptability is equipped to both the inner and outer loops simultaneously; 3) we demonstrate the efficacy of the proposed transfer learning strategy with experiments using a large set of clinical ultrasound images from prostate cancer patients, labelled by four different observers with varying experience and expertise; 4) the experiments show that using 20-30$\%$ of the expert labels is sufficient to fine-tune both the RL controller and the task predictor to achieve comparable performances to when they are trained using the full set of expert labels.

\section{Methods}

\subsection{Image quality assessment by task amenable data selection}\label{subsec:rl_iqa}

In this work, we follow the IQA formulation proposed by Saeed \textit{et al.} \cite{saeed_amenability}. There are two parametric functions, a task predictor $f(\cdot;w):\mathcal{X}\rightarrow\mathcal{Y}$ and a controller $h(\cdot;\theta):\mathcal{X}\rightarrow[0,1]$, with parameters $w$ and $\theta$, respectively. $\mathcal{X}$ and $\mathcal{Y}$ are the respective image and label domains with $\mathcal{P}_{XY}$ being the joint image-label distribution, with a density function $p(x,y)$.

The task predictor $f$ is optimised to predict labels, by minimising the loss function $L_f:\mathcal{Y}\times\mathcal{Y}\rightarrow\mathbb{R}_{\geq0}$ using sampled data:
\begin{align}
    \min_w \mathbb{E}_{(x,y)\sim\mathcal{P}_{XY}^h}[L_f(f(x;w), y)],
\end{align}
where $\mathcal{P}_{XY}^h$ is the controller-selected joint image-label distribution, with density function $p^h(x,y) \propto p(x,y)h(x;\theta)$.

The controller $h$ is optimised to measure image quality (task amenability), by minimising the metric function $L_h:\mathcal{Y}\times\mathcal{Y}\rightarrow\mathbb{R}_{\geq0}$:
\begin{align}
    \min_\theta \mathbb{E}_{(x,y)\sim\mathcal{P}_{XY}^h}[L_h(f(x;w), y)]
\end{align}
where $L_h$ is in general a non-differentiable metric computed on the validation set, and different to $L_f$.

The optimisation is performed using reinforcement learning, where the environment consists of the training set from $\mathcal{P}_{XY}$ and the task predictor $f(\cdot;w)$; the agent is the controller $h(\cdot;\theta)$ whose action is sample selection $a_t=\{a_{i,t}\}_{i=1}^B\in\{0,1\}^B$, based on the predicted quality scores $\{h(x_i;\theta)\}_{i=1}^B$, from a mini-batch of training samples $\mathcal{B}_t=\{(x_i,y_i)\}_{i=1}^B$; and the reward is the task predictor performance on a validation set from the same distribution $\mathcal{P}_{XY}$, which is computed after training, for a fixed number of steps, using the selected samples. In this work we use the reward formulation, from \cite{saeed_amenability}, which does not require human task amenability labels, and weights the validation set using controller predictions. $R_t$ is thus the reward which is a weighted sum of validation set performance.

\begin{figure}
    \centering
    \includegraphics[width=\textwidth]{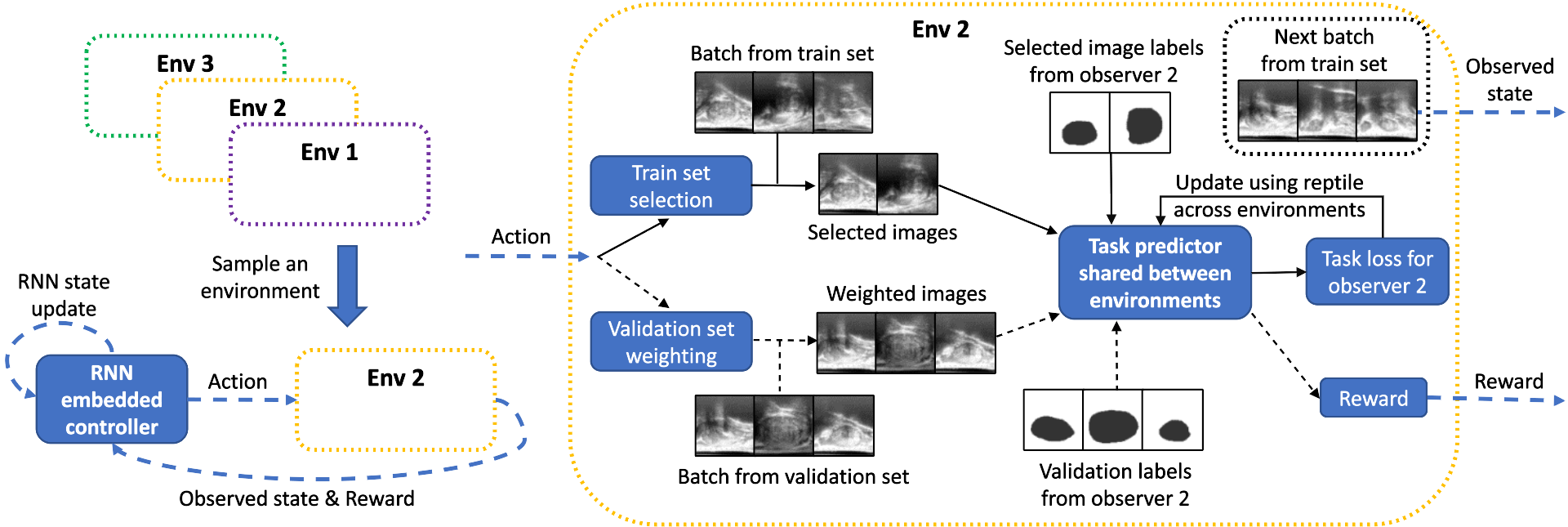}
    \caption{An overview of the proposed meta-RL framework for training the task predictor and the RNN-embedded controller (the IQA agent).}
    \label{fig:schematic}
\end{figure}

\subsection{Meta-reinforcement learning with different labels}\label{subsec:iqa_label}

In this section, we consider multiple label distributions $\{\mathcal{P}_{Y|X}^{k}\}_{k=1}^K$, such that each sample $x$ has multiple labels $\{y^k\}_{k=1}^K$. The joint distributions are thereby $\mathcal{P}_{XY}^{k}=\mathcal{P}_{X}\mathcal{P}_{Y|X}^{k}$ for $k=1,\ldots,K$.
Each distribution $\mathcal{P}_{XY}^{k}$ forms an RL environment with an MDP $M_k$. These MDPs are assumed to be sampled from the same MDP distribution $\mathcal{P}_M$, i.e. $M_k\sim\mathcal{P}_M$. The task predictors $f(\cdot;w)$ and controller $h(\cdot;\theta)$ are both shared across different environments.

We adopt the meta-RL formulation \cite{duan_metarl, wang_metarl} for reinforcement learning across multiple environments. Given a set of MDPs $\{M_k\}_{k=1}^K$, a \textit{trial} is defined as multiple episodes with a sampled MDP $M_k$. The meta-RL agent learns across multiple trials by sampling $M_k\sim\mathcal{P}_M$. Different from the RL with one single fixed environment, at time $t+1$, the meta-RL agent $h$ takes the action $a_t$, raw reward $r_t$, and termination flag $d_t$ at the previous time step in addition to the observed current state $s_{t+1}$. Note that for per-sample operation $r_t=R_t$ at the episode end, and zero otherwise, similar to sparse reward formulations in \cite{duan_metarl, wang_metarl}. 
Denote the input tuple as $\tau_{t+1}=(s_{t+1},a_t,r_t,d_t)$, thereby $h(\cdot;\theta)$ is now defined with a space of $\mathcal{X}\times[0,1]\times\mathbb{R}\times\{0,1\}$.

In this work, the meta-RL agent adopts a recurrent neural network (RNN) with internal memory shared across episodes in the same trial. Importantly, the internal memory is reset when a trial finishes, i.e. before another environment is sampled. This mechanism allows test-time adaptability, even with fixed weights \cite{cotter_rnn, santoro_rnn, younger_rnn, hochreiter_rnn, prokhorov_rnn}, and thereby transfers knowledge between environments \cite{duan_metarl, wang_metarl, botvinick_metarl}. This is due to the RNN making the controller a function of the history leading up to a sample such that changing history can influence the action for that sample.
The full algorithm is described in Algorithm \ref{algo:meta_iqa}, with details for configuring episodic mini-batches and meta-loop trials. An overview is also presented in Fig. \ref{fig:schematic}. In our implementation, proximal policy optimisation (PPO)~\cite{schulman_ppo} was used to train the controller. The task predictor employs the Reptile scheme \cite{nichol_reptile} to allow potential data efficiency benefit for adapting to different observer labels. The predictor is updated in two steps: 1) update starting weights $w_{t+1}$ of predictor $f(\cdot;w_{t+1})$\ to $w_{t+1,\text{new}}$, using gradient descent based on $\mathcal{B}_{t,\text{selected}}$; 2) update weights using $w_{t+1} \leftarrow w_{t+1} + \epsilon(w_{t+1, \text{new}}-w_{t+1})$ where $\epsilon$ is 1.0 initially and is linearly annealed to 0.0 as trial iterates. It is worth noting that the IQA algorithm from \cite{saeed_amenability} can be considered a special case of our proposed method with only one environment.

After training using the scheme described in Algorithm \ref{algo:meta_iqa}, the adaptation stage, for both the controller and task-predictor, can be performed on a single MDP of interest $M_a\sim\mathcal{P}_M$, where $M_a$ is the environment which we would like to adapt to. If multiple iterations of the outer loop are required, the internal state of the controller is only reset on the first iteration. The controller weights remain fixed; adaptability is a result of updating internal state.

\begin{algorithm}[!ht]
    \SetAlgoLined
    \KwData{Multiple MDPs $M_k\sim\mathcal{P}_M$.}
    \KwResult{Task predictor $f(\cdot;w)$ and controller $h(\cdot;\theta)$.}
    \BlankLine
    \While{not converged}{
        Sample an MDP $M_k\sim\mathcal{P}_M$\;
        Reset the internal state of controller $h$\;
            \For{Each episode in all episodes}{
                \For{$t\leftarrow 1$ \KwTo $T$}{
                    Sample a training mini-batch $\mathcal{B}_{t}=\{(x_{i,t},y_{i,t})\}_{i=1}^{B}$\;
                    Compute selection probabilities $\{h_{i,t}\}_{i=1}^B=\{h(\tau_{i,t};\theta_t)\}_{i=1}^B$\;
                    Sample actions $a_{t}=\{a_{i,t}\}_{i=1}^B$ w.r.t. $a_{i,t}\sim\text{Bernoulli}(h_{i,t})$\;
                    Select samples $\mathcal{B}_{t,\text{selected}}$ from $\mathcal{B}_{t}$\;
                    Update predictor $f(\cdot;w_t)$\ with $\mathcal{B}_{t,\text{selected}}$ using Reptile\;
                    Compute reward $R_{t}$\;
                }
                Collect one episode $\{\mathcal{B}_t,a_t,R_t\}_{t=1}^T$\;
        
                Update controller $h(\cdot;\theta)$ using the RL algorithm PPO\;
            }
        }
\caption{Adaptable image quality assessment by task amenability}\label{algo:meta_iqa}
\end{algorithm}

\section{Experiments}\label{sec:exp}

In this work we use 6644 2D ultrasound images from 249 prostate cancer patients. During the early stages of ultrasound-guided biopsy procedures, images were acquired using a transperineal ultrasound probe (C41L47RP, HI-VISION Preirus, Hitachi Medical Systems Europe) as part of SmartTarget: THERAPY and SmartTarget: BIOPSY clinical trials (clinicaltrials.gov identifiers NCT02290561 and NCT02341677 respectively). Images from each subject initially consisted of 50-120 frames. For feasibility of manual labelling, frames were sampled at four-degree intervals where relative rotation angles were tracked using a digital transperineal stepper (D\&K Technologies GmbH, Barum, Germany). The resulting 6644 2D ultrasound images were randomly split, at the patient level, into training, validation and holdout sets, with 4429, 1092 and 1123 images from 174, 37 and 38 subjects, respectively.

Three sets of task label $\{L_i\}_{i=1}^3$ were collected from three trained biomedical engineering researchers. These individually-labelled are referred to as ``non-expert'' label sets for brevity. In addition, the fourth set of ``expert'' labels $L_*$ was curated by a urologist, first carefully reviewing a reference set of consensus labels and then editing them as deemed necessary. For all label sets, each image has both a binary label indicating prostate presence for classification and a binary mask of the prostate gland for segmentation.

The task predictor algorithms used for the two tested applications are the same as \cite{saeed_amenability}. For classification, AlexNet \cite{alexnet_class} was used with a cross-entropy loss and a reward based on classification accuracy. For segmentation, U-Net \cite{unet_seg} was used with a pixel-wise cross-entropy loss and a reward based on mean binary Dice score. The controller had a three-layer convolutional encoder, before feeding the encoded features to an RNN with a stacked-LSTM architecture, as described in \cite{wang_metarl}. Experimental results are reported for empirically configured networks and default hyper-parameter values remain unchanged unless specified. 

The following three different IQA models were trained and compared.
\begin{itemize}
    \item \textit{Baseline}: Trained with all training and validation data using only the high-quality expert labels $L_*$. That is, only one ``expert-labelled'' environment in training, establishing a reference for achievable IQA system performance.
    \item \textit{Meta-RL}: The proposed model that was first trained with training and validation data using the non-expert labels $\{L_i\}_{i=1}^3$ as three different environments. Both the task predictor and the controller were subsequently adapted with $k\times100\%$ training and validation data using the expert labels $L_*$.
    \item \textit{Meta-RL Variant}: For comparison, a basic implementation of transfer learning. The model was first trained with all training and validation data using the shuffled non-expert labels $\{L_i\}_{i=1}^3$ as one single environment, i.e. without considering different environment-specific trials, and the Reptile update for optimising the task predictor reduced to standard gradient descent. Adaptation was done with $k\times100\%$ training and validation data using the expert labels $L_*$. The internal state of RNN was not reset before fine-tuning.
\end{itemize}

We evaluate the IQA models jointly with the task predictors using task performance, which serves as both a direct evaluation of the task-predictor and an indirect evaluation of the IQA agent by its task amenability definition. We report mean accuracy (Acc.) and mean binary Dice score (Dice) on the holdout set using expert labels for classification and segmentation, respectively. These measures are averaged over all 2D slices in the holdout set. Where controller selection is used, the metric is computed over the selected samples only. Samples are selected by rejecting the subset with the lowest controller predicted values, with the specified rejection ratios. Standard deviation (St.D) is reported to measure inter-patient variance, with which, paired t-test results with a significance level of 5\% are reported when any comparison is made. We evaluate the models for varying $k$-values, where $k$ is the ratio of expert-labelled samples used for adaptation ($k\times100\%$ samples used).

\section{Results}\label{sec:results}

\begin{table}[!ht]
\centering
\caption{Comparison of holdout set results with a rejection ratio set to 5\%}
\begin{tabular}{|c|c|c|c|}
\cline{1-4}
\multicolumn{2}{|c|}{Tasks} & Prostate Classification (Acc.) & Prostate Segmentation (Dice)\\
\cline{1-4}
IQA Methods & k & Mean $\pm$ St.D. & Mean $\pm$ St.D.\\
\cline{1-4}
Baseline & N/A & 0.932 $\pm$ 0.011 & 0.894 $\pm$ 0.016\\
\cline{1-4}
\multirow{6}{*}{Meta-RL} & 0.5 & 0.936 $\pm$ 0.012 & 0.892 $\pm$ 0.018\\
\cline{2-4}
& 0.4 & 0.929 $\pm$ 0.016 & 0.886 $\pm$ 0.014 \\
\cline{2-4}
& 0.3 & 0.926 $\pm$ 0.010 & 0.888 $\pm$ 0.020\\
\cline{2-4}
& 0.2 & 0.925 $\pm$ 0.017 & 0.873 $\pm$ 0.017\\
\cline{2-4}
& 0.1 & 0.911 $\pm$ 0.012 & 0.863 $\pm$ 0.020\\
\cline{2-4}
& 0.0 & 0.908 $\pm$ 0.010 & 0.857 $\pm$ 0.018\\
\cline{1-4}
\multirow{6}{*}{Meta-RL Variant} & 0.5 & 0.931 $\pm$ 0.015 & 0.884 $\pm$ 0.016\\
\cline{2-4}
& 0.4 & 0.920 $\pm$ 0.010 & 0.882 $\pm$ 0.021\\
\cline{2-4}
& 0.3 & 0.919 $\pm$ 0.013 & 0.882 $\pm$ 0.015\\
\cline{2-4}
& 0.2 & 0.916 $\pm$ 0.014 & 0.860 $\pm$ 0.014\\
\cline{2-4}
& 0.1 & 0.905 $\pm$ 0.014 & 0.858 $\pm$ 0.021\\
\cline{2-4}
& 0.0 & 0.896 $\pm$ 0.016 & 0.849 $\pm$ 0.017\\
\hline
\end{tabular}
\label{tab:res}
\end{table}

The proposed meta-training took, on average, approximately 48 hours and the meta-testing (model fine-tuning) took 1-2 hours on a single Nvidia Quadro P5000 GPU. This result reflects the design of the proposed adaptation strategy for data efficiency and, arguably, also for computational efficiency.

\begin{figure}[!ht]
  \centering
  \subfloat[Prostate presence classification task]{\includegraphics[width=0.49\textwidth, height=3.5cm]{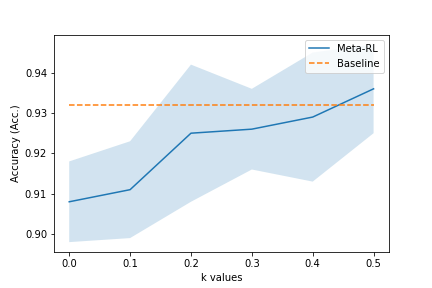}\label{fig:acc_ratio}}
  \hfill
  \subfloat[Prostate segmentation task]{\includegraphics[width=0.49\textwidth, height=3.5cm]{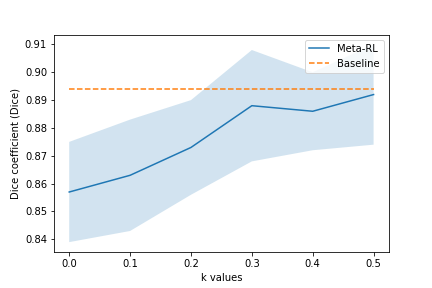}\label{fig:dice_ratio}}
  \caption{Task performance (in respective Acc. and Dice metrics) against the $k$ values with rejection ratio set to 5\%.}
  \label{fig:res:perf_h2_ratio}
\end{figure}

Performance of the IQA models, in terms of Acc. and Dice, are summarised in Table \ref{tab:res} and plotted in Fig. \ref{fig:res:perf_h2_ratio} against varying $k$ values. In the prostate presence classification task, no statistical significance was found between the baseline and meta-RL for $k$ values from $0.5$ to $0.2$ ($p$-values ranged from 0.10 to 0.23). However, meta-RL performance for low $k$ values, $k=0.1$ or $0.0$, was significantly lower than that of the baseline ($p<0.01$ for both). In the prostate segmentation task, no statistical significance was found between the two, for $k$-values from $0.5$ to $0.3$ ($p$-values ranged from 0.07 to 0.17), but a significantly lower performance was found for meta-RL for low $k$ values from $0.2$ to $0.0$ ($p$<0.01 for all). 

For the ablation study comparing meta-RL to the meta-RL variant, the proposed meta-RL framework generally outperformed the meta-RL variant for the same $k$ values, for both tested target tasks, as detailed in Table \ref{tab:res}. For classification, we report a statistically significant difference between the two, for the same $k$ values from 0.0 to 0.4 (p<0.01 for all), while no significance was found when the $k$ increased to 0.5 ($p$=0.06). For segmentation, superior performance from the proposed meta-RL was statistically significant for all $k$ values ($p$<0.03 for all). From an ablation study, with and without the Reptile scheme for updating task predictors, the Reptile-omitted meta-RL classification and segmentation tasks achieved Acc.=$0.901\pm0.013$ and Dice=$0.851\pm0.013$, respectively, when $k=0$. The improvement, when using the Reptile scheme, was statistically significant with $p$<0.01 for both, but no significant difference was found for other $k$ values.

\begin{figure}[!ht]
  \centering
  \subfloat[Prostate classification task]{\includegraphics[width=0.4\textwidth, height=6.25cm]{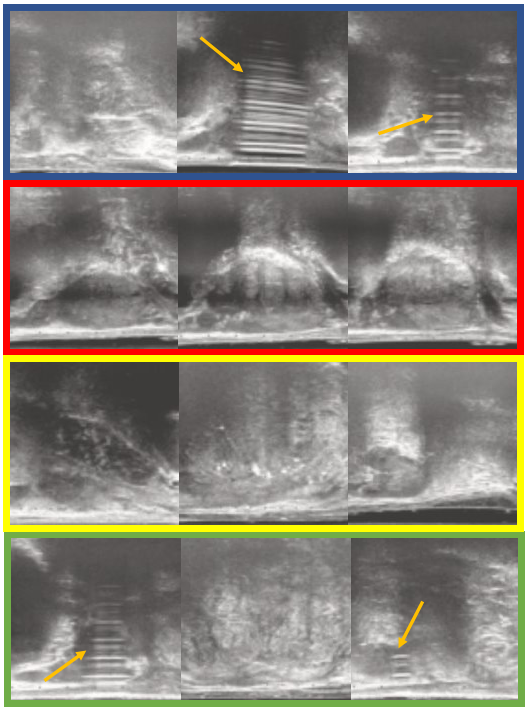}\label{fig:samples_class}}
  \hfill
  \subfloat[Prostate segmentation task]{\includegraphics[width=0.4\textwidth, height=6.25cm]{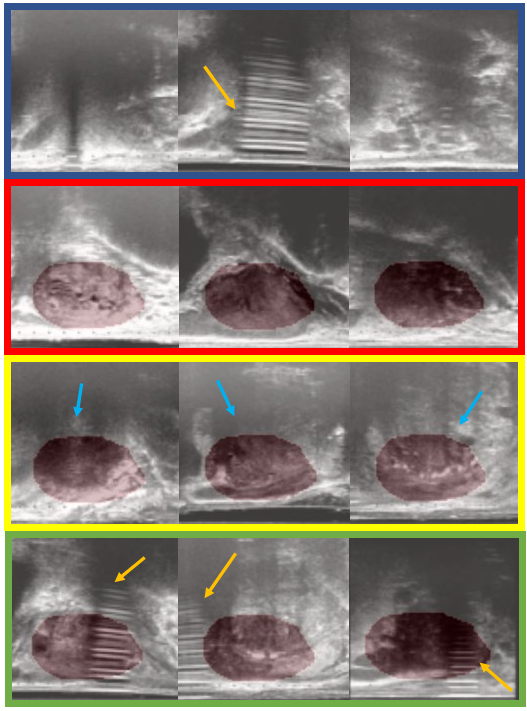}\label{fig:samples_seg}}
  \caption{Examples of controller selected and rejected images (rejection ratio=5\%) for both tasks. \textbf{Blue:} rejected samples; \textbf{Red:} selected samples; \textbf{Yellow:} rejected samples despite no apparent artefacts or severe noise; \textbf{Green:} selected samples despite present artefacts or low contrast. \textbf{Orange arrows:} visible artefacts; \textbf{Cyan arrows:} regions where gland boundary delineation may be challenging.}
  \label{fig:res:samples}
\end{figure}

Fig. \ref{fig:res:samples} provides visual examples of selection decisions by the adapted IQA agent. With 5$\%$ rejection ratio, all these rejected examples seem visually challenging for respective classification and segmentation tasks, and rejecting these examples improved performances of the simultaneously learned task predictors. 

\section{Discussion and Conclusion}

Based on results reported in Sect. \ref{sec:results}, for the tested ultrasound guidance application, the proposed adaptation strategy allows for the IQA agent and task predictor to be adapted using as few as 1087 and 1634 expert-labelled images from 42 and 63 subjects (training and validation sets), for classification and segmentation, respectively. Compared with a total of 5521 expert-labelled images from 211 subjects that were required to train the baseline, this is a substantial reduction, to 20-30\%, in the required quantity of high-quality and often expensive expert-labelled data. The proposed model also used non-expert labels for training but these may be used for different IQA definitions, further economic analysis is beyond the scope of this work. An adaptable IQA algorithm has been presented, which can be efficiently adapted with new labelled data. The proposed algorithm may have general applicability to alleviate demand for large quantities of training data, for example, for other imaging protocols or target tasks.

\section*{Acknowledgements}

This work is supported by the Wellcome/EPSRC Centre for Interventional and Surgical Sciences [203145Z/16/Z], the CRUK International Alliance for Cancer Early Detection (ACED) [C28070/A30912; C73666/A31378], EPSRC CDT in i4health [EP/S021930/1], the Departments of Radiology and Urology, Stanford University, an MRC Clinical Research Training Fellowship [MR/S005897/1] (VS), the Natural Sciences and Engineering Research Council of Canada Postgraduate Scholarships-Doctoral Program (ZMCB), the University College London Overseas and Graduate Research Scholarships (ZMCB), GE Blue Sky Award (MR), and the generous philanthropic support of our patients (GAS). Previous support from the European Association of Cancer Research [2018 Travel Fellowship] (VS) and the Alan Turing Institute [EPSRC grant EP/N510129/1] (VS) is also acknowledged.

\printbibliography

\end{document}